\documentclass[sigconf,nonacm]{acmart}

\usepackage[utf8]{inputenc}                 
\usepackage[font=small,labelfont=bf]{caption} 
\usepackage{booktabs, siunitx, xcolor}
\usepackage{hyperref}
\usepackage{booktabs}
\usepackage{enumitem}
\usepackage{indentfirst}
\usepackage{xspace}
\usepackage{comment}
\usepackage{amsmath}
\usepackage{amsfonts}
\usepackage{textcomp}
\usepackage{url} 
\usepackage{graphicx}    
\usepackage{multirow}
\usepackage{cleveref}
\usepackage{float}
 
\newcommand{\name}{\textbf{\texttt{AutoEDA}}\xspace}

\usepackage[most]{tcolorbox}
\tcbuselibrary{listings, breakable}
\usepackage{listings}
\usepackage{pifont}

\newtcolorbox{bluebox}[2][]{
  enhanced jigsaw,
  breakable,
  colback=blue!2!white,
  colframe=blue!50!black,
  boxrule=0.5pt,
  arc=2mm,
  left=6pt, right=6pt, top=0pt, bottom=0pt,
  width=\linewidth, 
  title={#2},
  fonttitle=\small\bfseries\ttfamily,
  coltitle=blue!40!black,
  colbacktitle=blue!4!white,
  boxed title style={
    sharp corners,
    boxrule=0.5pt,
    colframe=blue!50!black,
    left=2pt, right=2pt, top=0pt, bottom=0pt
  },
  attach boxed title to top left={xshift=6pt, yshift*=-3mm},
  borderline west={1.6mm}{0pt}{blue!60!black},
  before skip=2pt,
  after skip=2pt,
  #1
}

\setcopyright{acmlicensed}
\copyrightyear{2026}
\acmYear{2026}
\acmDOI{XXXXXXX.XXXXXXX}
\acmConference[ACM/IEEE Design Automation Conference]{}{JULY 26--29, 2026}{LONG BEACH, CA}
\acmISBN{978-1-4503-XXXX-X/2018/06}

\begin{document}

\title{AutoEDA: Enabling EDA Flow Automation through Microservice-Based LLM Agents}

\author{Yiyi Lu}
\authornote{Equal contribution. Work done during an internship at Duke University.}
\affiliation{%
  \institution{Duke University}
  \country{USA}
}

\author{Hoi Ian Au}
\authornotemark[1]
\affiliation{%
  \institution{Duke University}
  \country{USA}
}

\author{Junyao Zhang}
\affiliation{%
  \institution{Duke University}
  \country{USA}
}

\author{Jingyu Pan}
\affiliation{%
  \institution{Duke University}
  \country{USA}
}

\author{Guanglei Zhou}
\affiliation{%
  \institution{Duke University}
  \country{USA}
}

\author{Yiting Wang}
\affiliation{%
  \institution{University of Maryland, College Park}
  \country{USA}
}

\author{Jingwei Sun}
\affiliation{%
  \institution{University of Florida}
  \country{USA}
}

\author{Ang Li}
\affiliation{%
  \institution{University of Maryland, College Park}
  \country{USA}
}

\author{Jianyi Zhang}
\affiliation{%
  \institution{Duke University}
  \country{USA}
}

\author{Hai Li}
\affiliation{%
  \institution{Duke University}
  \country{USA}
}

\author{Yiran Chen}
\affiliation{%
  \institution{Duke University}
  \country{USA}
}

\begin{abstract}
Electronic Design Automation (EDA) remains heavily reliant on tool command language (Tcl) scripting to drive complex RTL-to-GDSII flows. This scripting-based paradigm is labor-intensive, error-prone, and difficult to scale across large design projects. Recent advances in large language models (LLMs) suggest a new paradigm of natural language–driven automation. However, existing EDA efforts remain limited and face key challenges, including the absence of standardized interaction protocols and dependence on external APIs that introduce privacy risks.

We present \name, a framework that leverages the Model Context Protocol (MCP) to enable end-to-end natural language control of RTL-to-GDSII design flows. \name introduces MCP-based servers for task decomposition, tool selection, and automated error handling, ensuring robust interaction between LLM agents and EDA tools. To enhance reliability and confidentiality, we integrate locally fine-tuned LLM agents. We further contribute a benchmark generation pipeline for diverse EDA scenarios and extend CodeBLEU with Tcl-specific enhancements for domain-aware evaluation. Together, these contributions establish a comprehensive framework for LLM-driven EDA automation, bridging natural language interfaces with modern chip design flows. Empirical results show that \name achieves up to 9.9$\times$ higher accuracy than naïve approaches while reducing token usage by approximately 97\% compared to in-context learning.

\end{abstract}

\maketitle

\section{Introduction} \label{sec:intro}

Contemporary Electronic Design Automation (EDA) encompasses a suite of software tools used for designing, analyzing, and verifying integrated circuits (ICs). Among the most complex components of EDA flow are the synthesis and physical design stages (RTL-to-GDSII), which involve numerous procedures and highly configurable parameters \cite{synopsys_dc_user_guide, synopsys_dc_methodology, cadence_innovus_user_guide}. Traditionally, engineers interact with EDA tools by writing custom scripts, typically in the Tool Command Language (Tcl), to specify constraints, control tool behavior, and coordinate multi-stage execution \cite{chen2001scripting, innovus_methodology_paper}. However, this scripting-based workflow is labor-intensive, error-prone, and difficult to scale, especially in large and heterogeneous projects. Moreover, maintaining compatibility across different vendor ecosystems further complicates script development and reuse, making the traditional approach inefficient in modern EDA environments.

\begin{figure}[t]
  \centering
  \includegraphics[width=0.9\linewidth]{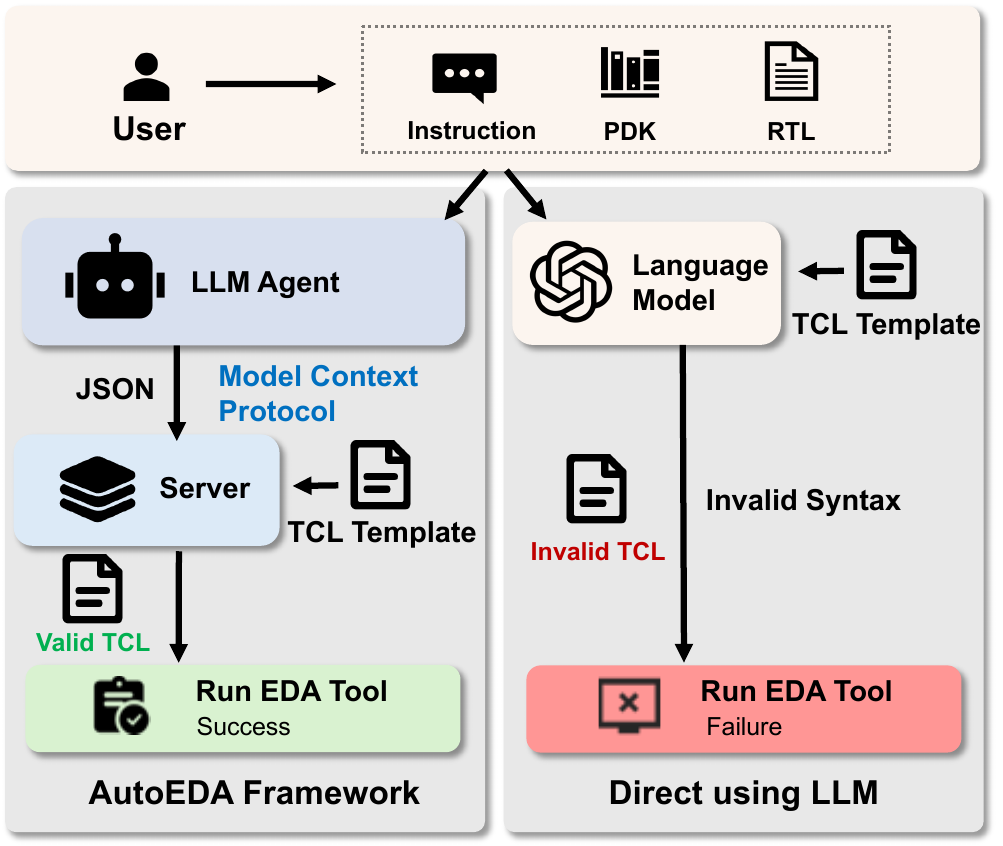}
  \caption{
  Comparison between the \name and the direct LLM baseline on RTL-to-GDSII flow automation. \name leverages LLM agent as client together with Model Context Protocol servers to generate optimized TCL scripts, enabling reliable execution in EDA tools and producing valid results. 
  }
  \label{fig:teaser}
  \vspace{-10pt}
  \Description{}
\end{figure}

Recent advances in Large Language Models (LLMs) have demonstrated remarkable capabilities in tool invocation and workflow orchestration in a variety of domains \cite{guo2025survey}. Frameworks like ToolFormer \cite{toolformer}, xLAM \cite{xlam}, and ToolACE \cite{toolace} illustrate that LLMs can effectively interpret tool functionalities, generate correct API calls, and coordinate complex multi-step processes through natural language interfaces. 
These developments highlight the potential of LLMs to bridge human intent with sophisticated tool ecosystems, suggesting a promising new paradigm for EDA automation.
Recent studies have explored applying LLMs to EDA. Works such as RTL code generation \cite{rtlcoder, rtllm, chipchat}, demonstrating the feasibility of automated hardware description language synthesis. However, examples of language-driven flow control remain limited, and existing solutions are often unstable, lacking standardized workflows and relying on complex back-end integration \cite{chateda}.

As illustrated in the right panel of \Cref{fig:teaser}, relying on an LLM to generate Tcl commands frequently results in syntactic errors and invalid scripts that fail in downstream tools. These shortcomings reveal four challenges for deployment of LLM-based EDA automation:
First, there is no standardized interaction protocol for LLM–EDA tool interaction, forcing reliance on bespoke interfaces that are difficult to generalize across design contexts and tools.
Second, privacy and reliability risks arise from dependence on external API services, which pose intellectual property risks for semiconductor companies dealing with proprietary designs. Moreover, these models often produce syntactically incorrect or semantically invalid Tcl scripts due to limited domain-specific training, causing tool execution failures.
Third, the lack of standardized benchmarks makes systematic evaluation and fair comparison across approaches impossible.
Finally, evaluation metrics remain inadequate, since conventional code quality measures fail to capture EDA-specific requirements. Such metrics cannot determine whether generated scripts include valid tool commands or use parameters correctly, leaving the practical utility of automated scripts insufficiently assessed.

To address these limitations, we present \name, a comprehensive framework that utilizes the Model Context Protocol (MCP) \cite{mcp} to enable end-to-end natural language control of RTL-to-GDSII design flows. Our approach systematically tackles the identified shortcomings through four key innovations: (1) we implement an MCP-based server that supports task decomposition, tool selection, and automated bug handling thereby providing a robust interaction layer between LLM agent and EDA tools; (2) we employ locally fine-tuned agent that eliminates dependencies on external LLM service while enhancing performance, and safeguards design confidentiality; (3) we introduce a dataset generation pipeline that covers diverse EDA scenarios; (4) we extend CodeBLEU to support Tcl script evaluation, enabling more precise assessment of code quality with respect to both syntactic correctness and EDA-specific semantics. This paper makes the following major contributions:
\begin{itemize}[leftmargin=*]
    \item We propose \name, a holistic framework built on Model Context Protocol. It integrates locally fine-tuned LLM agent with a microservice-based backend, enabling natural language control of RTL-to-GDSII design flows.
    \item \name is supported by microservice-based \textbf{MCP servers} that facilitate task decomposition into four major sub-flows (synthesis, placement, clock tree synthesis, and routing), while enabling automated syntax handling.
    \item We introduce a \textbf{data generation pipeline} covering various EDA scenarios and extend \textbf{CodeBLEU-Tcl}, a domain-aware metric designed to assess both the syntactic correctness and semantic validity of generated scripts.
\end{itemize}

\section{Preliminaries} \label{sec:bg}

\subsection{LLM Agents and the Model Context Protocol}
\begin{figure}[t]
  \centering
  \includegraphics[width=0.9\linewidth]{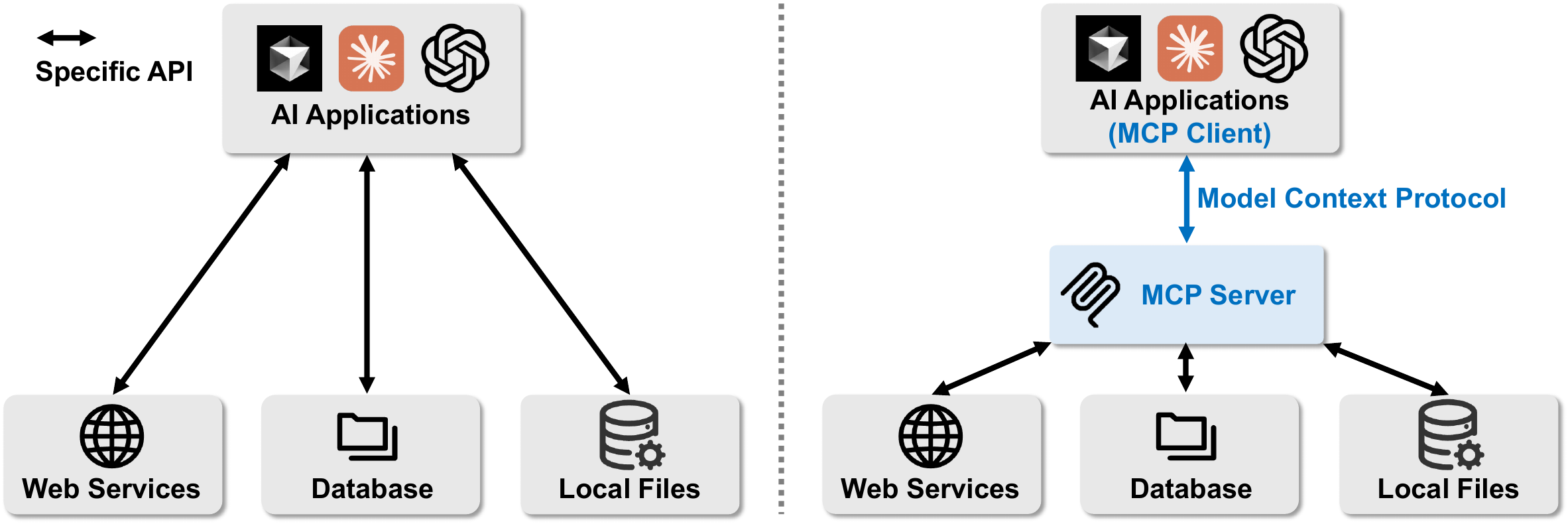}
  \caption{\textbf{MCP server roles in tool invocation.} Direct tool invocation on the left versus invocation with MCP servers on the right.}
  \label{fig:mcp}
  \vspace{-10pt}
  \Description{}
\end{figure}

LLM agents can be viewed as autonomous software entities that leverage an LLM as the core reasoning mechanism to perceive their environment, select actions, and follow goals \cite{wang2024survey}. Unlike rule-based automation, they exhibit autonomy, social ability, reactivity, and proactiveness\cite{wooldridge2009introduction}. Modern implementations extend base models with modules for memory, task planning, tool use, and real-world execution \cite{guo2024large}. A key milestone was the introduction of structured function calling in 2023, enabling models to generate JSON payloads conforming to predefined API schemas, thereby tightly coupling natural-language intent with executable actions \cite{toolformer, wang2023voyager, openai2023function}.

To ensure reliable and reusable tool interactions, Anthropic’s Model Context Protocol (MCP) defines a JSON-RPC–based interface layer between models and external resources \cite{singh2025survey}. As shown in \Cref{fig:mcp}, MCP standardizes client–server separation, where clients request capabilities and servers expose tools, resources, and prompts. \cite{ehtesham2025survey, anthropic2024mcp,ray2025survey}. The spec standardizes four primitives: tools (model-controlled API calls), resources (application-curated data objects), prompts (reusable instruction templates), and sampling (outsourcing generation to another model or server) to enable consistent multi-step orchestration across diverse ecosystems \cite{dragoni2017microservices, newman2015microservices}.

\subsection{VLSI Design Flow}

\begin{figure*}[t]
  \centering
  \includegraphics[width=0.8\textwidth]{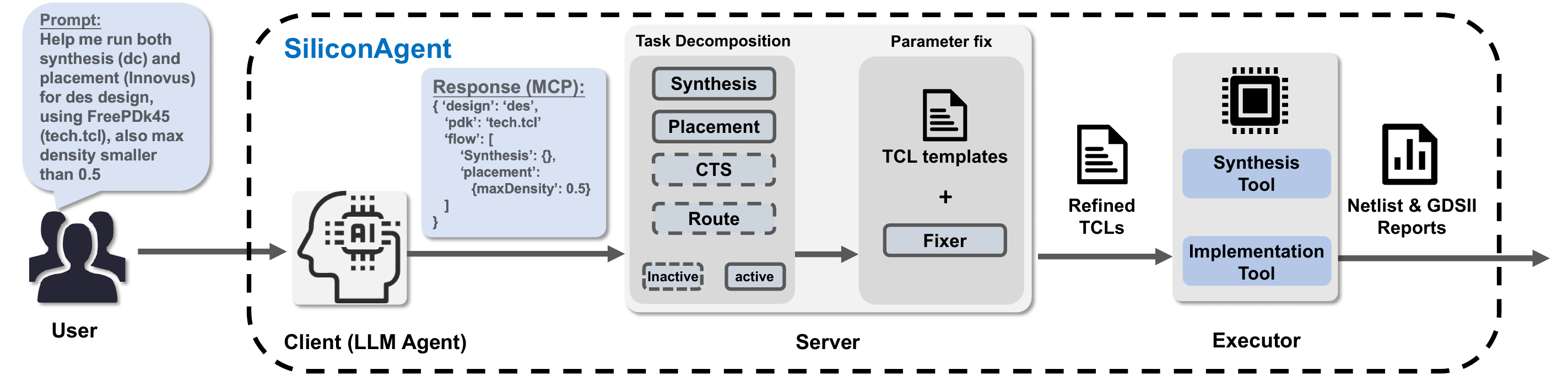}
  \caption{\textbf{\name overview.} 
Natural-language intents are first transformed into structured IR by the client, then validated and rendered into Tcl scripts by the MCP server, before execution in commercial EDA tools.}
  \label{fig:overview}
  \vspace{-10pt}
  \Description{}
\end{figure*}
The implementation of VLSI circuits follows a structured and multi-stage process that transforms high-level specifications into manufacturable chip layout\cite{gao2023review, zhong2023llm4eda}. A standard flow typically includes four key stages: specification definition, RTL design, logic synthesis, and physical design. Among these, logic synthesis \cite{synthesis} and physical design \cite{physical_design} are the most complex and domain-specific. 

Logic synthesis translates RTL code into an intermediate representation (IR), such as the AND-Inverter Graph (AIG) in the ABC tool \cite{abc}, optimizes it to minimize depth and delay, and maps the result onto standard cell libraries provided by foundries. The outcome is a gate-level netlist describing the circuit as interconnected logic gates \cite{itrs2024reports}.
Physical design then transforms the above gate-level netlist into a layout through sequential steps. Floorplanning organizes functional blocks to balance performance and area. Placement assigns exact gate locations while reducing wire length and congestion. Next, Clock tree synthesis (CTS) builds a synchronized distribution network to meet timing. Finally, routing establishes signal and power connections using metal interconnects.

\section{Methodology} \label{sec:mcp}

\subsection{Overview}
\Cref{fig:overview} outlines a systematic approach for \name which translates natural-language design intents into production-ready RTL-to-GDSII flows. The framework is organized into three layers: \emph{client}, \emph{server}, and \emph{executor}. Users first provide RTL designs, PDK, and describe goals in free text. The locally hosted, fine-tuned LLM agent (client) converts this input into a typed, tool-agnostic specification that declares \emph{what} tasks to execute, and \emph{which} parameters to use. The MCP-based server validates, decomposes, and refines this specification before rendering it into executable Tcl scripts. These scripts are then executed by commercial or open-source EDA tools in the executor, producing artifacts and reports.

\begin{figure}[t]
  \centering
  \includegraphics[width=0.8\linewidth]{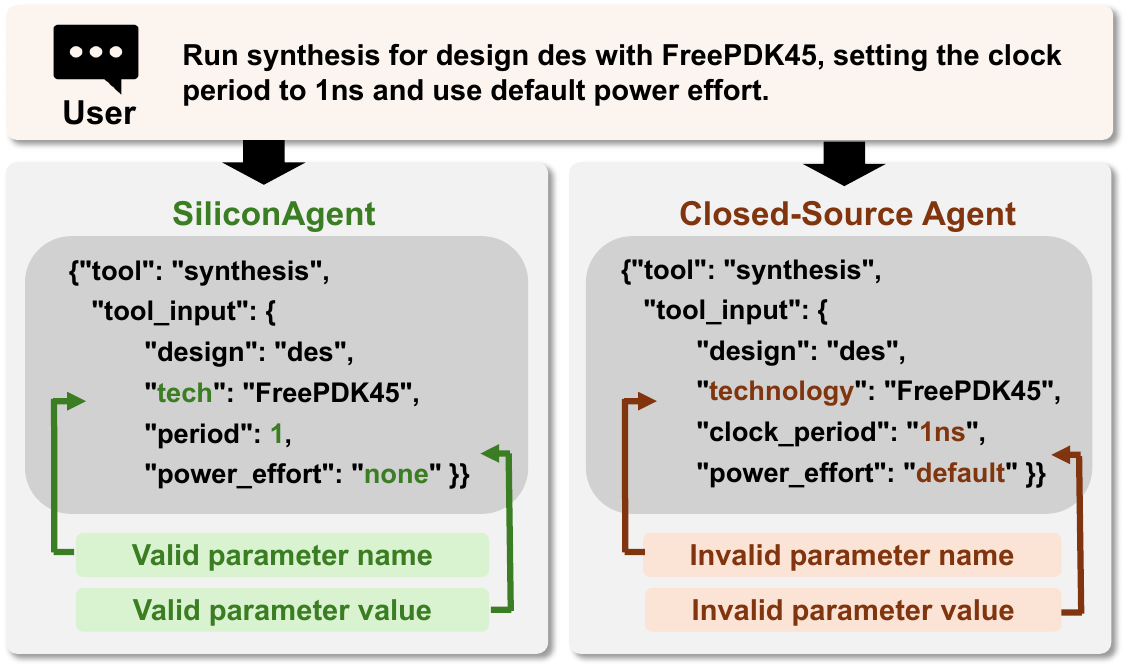}
  \caption{\textbf{Closed-source client error.}
  Closed-source models often produce invalid IRs that have incorrect field names or values.}
  \label{fig:client}
  \vspace{-10pt}
  \Description{}
\end{figure}
\subsection{Client} \label{sec:client}
The framework begins with the client, where a locally hosted language model interprets natural-language queries. As shown in \Cref{fig:teaser}, directly generating Tcl scripts often leads to syntactic and semantic errors. To address this, the client instead produces a structured intermediate representation (IR) based on the query:
\[
\text{Client}(\text{query}) \;\longrightarrow\; \{\text{tool},\,\text{params}\}
\]
The IR is a schema-driven JSON object with explicit types (numeric with units, categorical enums, booleans, lists), stage identifiers, and cross-stage constraints. This IR is then wrapped in Model Context Protocol format and passed to the server.  

Our initial client prototype was built with closed-source models, but two critical limitations were observed. First, \emph{Reliability}: despite careful prompt engineering, models often failed to produce valid IRs that conformed to the server’s schema. Errors such as incorrect field names or omission of essential parameters were common, as illustrated in \Cref{fig:client}. 
Second, \emph{Privacy}: reliance on external APIs required transferring design files and PDK assets outside the enterprise boundary, which is unacceptable in production environments.

To address these issues, \name employs a locally fine-tuned LLM agent as a client trained on benchmark data generated by the pipeline described in \Cref{sec:benchmark}. This fine-tuned client enforces schema compliance through grammar-constrained decoding, stage-aware exemplars, and rule-based post-processing, thereby ensuring both reliability and privacy in the front end.

\subsection{MCP Server}\label{sec:mcp_server}

The server acts as the authoritative mediator between the LLM client and commercial EDA tools. 
Its design follows a template–-method pattern with three layers of abstraction as follows:

\noindent\textbf{Task decomposition.} The server decomposes IR into standardized workflow phases corresponding to four major RTL-to-GDSII stages: synthesis, placement, CTS, and routing. Dependencies across stages are explicitly tracked, ensuring that later tasks (e.g., routing) are not executed before their prerequisites (e.g., placement). Each stage inherits from a base planner class that defines \texttt{precheck}, \texttt{inputs}, \texttt{actions}, and \texttt{outputs}, enabling modularity and extensibility.  

\noindent\textbf{Parameter validation.} 
Parameters are verified through a constraint-satisfaction mechanism. Invalid values are clipped to valid ranges (e.g., placement density capped at 1.0), missing parameters are inferred, and conflicting settings are automatically resolved. This improves robustness against both model and user introduced errors.  

\noindent\textbf{Template rendering.} The server collects the validated parameters and renders each task into a Tcl script. In this manner, we can avoid the command error and achieve a version-adaptive configuration.
The transformation can be expressed as
\[
\text{Server}(\text{templates},\,\text{fixed(params)})
   \;\longrightarrow\; \text{Tcl}\;\text{scripts},
\]
\noindent Compared to direct LLM-based generation, this layered design provides systematic error checking, consistent parameter handling, and tool-version awareness, thereby greatly improving reliability and success across heterogeneous EDA toolchains.

\subsection{Executor}
Once Tcl scripts are generated, the executor submits them to commercial or open-source EDA tools. Both synthesis (e.g., Synopsys DC) and implementation tools (e.g., Cadence Innovus) are supported. Logs and intermediate artifacts are continuously monitored, and results, including netlists, GDSII layouts, and reports, are returned to the user. In case of execution failure, the server engages its built-in fixer to diagnose logs and apply rule-based bounded repairs, ensuring robustness across heterogeneous toolchains.

\subsection{Benchmark Generation Pipeline}\label{sec:benchmark}

\begin{figure}[t]
  \centering
  \includegraphics[width=0.8\linewidth]{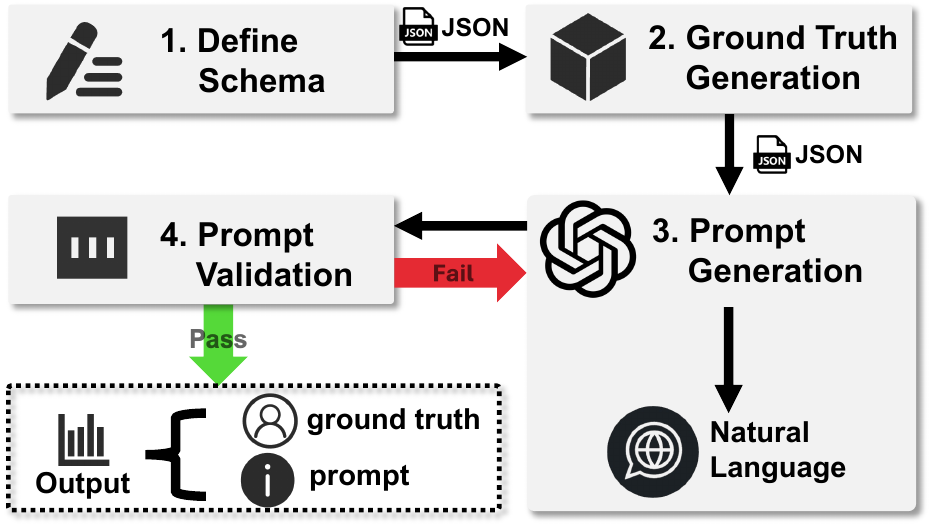}
  \caption{\textbf{Benchmark generation pipeline.}
  1) Define a schema of Tcl parameters with valid ranges/values.
  2) Randomly pick EDA flow steps and parameter settings from the schema. Multiple steps may be selected.
  3) LLM is used to convert the ground-truth into natural user instructions.
  4) Validate the generated prompt with rules.}
  \label{fig:benchmark}
  \vspace{-10pt}
  \Description{}
\end{figure}

Current EDA flow automation lacks standardized benchmarks for systematic evaluation. To address this gap, we design a bottom-up benchmark pipeline producing realistic prompts paired with ground-truth Tcl configurations, as illustrated in \Cref{fig:benchmark}. Inspired by tool-augmented dialog synthesis frameworks \cite{toolace}, our pipeline extends beyond static prompt–Tcl pairs by incorporating hierarchical parameter specification, adaptive sampling, and multi-agent dialog generation. The pipeline consists of the following steps:

\noindent\textbf{Schema Definition.}  
We first define a hierarchical parameter schema that specifies Tcl variables, valid ranges, and categorical options across the four stages introduced in \Cref{sec:mcp_server}.

\noindent\textbf{Ground-Truth Configuration Generation.}  
Configurations are sampled from this schema by selecting one or more stages and instantiating parameters from valid ranges or categories. To ensure coverage, we adopt an \emph{evolutionary adaptation strategy} \cite{toolace}
generating simple cases with minimal parameters and complex cases with additional constraints, perturbed values, or rare options. Each configuration is serialized into a structured JSON as ground truth.

\noindent\textbf{Prompt Generation.}
Each structured configuration is transformed into a natural-language query using LLM. To emulate diverse user behaviors, we generate stylistic variations, including directive commands, conversational queries, and code-like specifications. This ensures that benchmark captures the variability of real-world user inputs while remaining aligned with ground-truth configuration.

\noindent\textbf{Validation.}
All generated prompts undergo rule-based checks to ensure they reflect the intended design stages and parameter values. Invalid prompts are regenerated, guaranteeing semantic alignment between the natural-language input and its structured ground truth.

This pipeline yields a diverse dataset spanning different EDA stages, parameter granularities, and interaction modalities, providing a robust foundation for training and evaluating natural language–driven EDA automation systems.

\subsection{CodeBLEU-Tcl} \label{sec:codebleu}
Generic code evaluation metrics such as CodeBLEU \cite{codebleu} are designed for general-purpose languages (e.g., Python) and lack support for domain-specific scripting languages like Tcl. To quantify this limitation, we conducted a controlled experiment with 12 EDA script variants spanning different stages. Each variant was implemented in two forms: \textit{Standard Tcl}, which adheres to EDA conventions, and \textit{Pythonic Tcl}, which mimics EDA commands using Python-like syntax that cannot be executed by EDA tools.

\begin{figure}[t]
    \centering
    \includegraphics[width=0.9\linewidth]{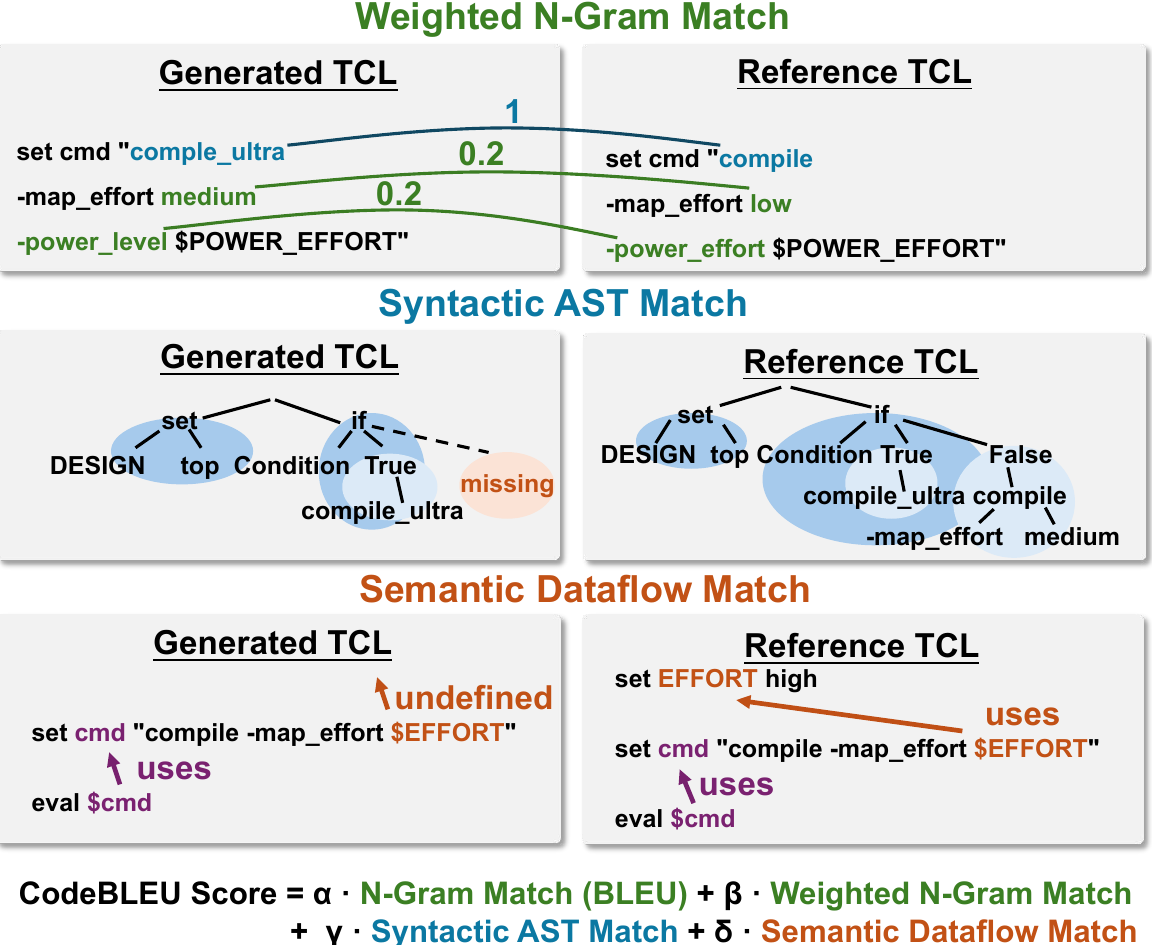}
    \caption{\textbf{CodeBLEU-Tcl.} CodeBLEU-Tcl is a combination of n-gram match, weighted n-gram match, syntactic AST match, and semantic dataflow match with EDA-specific weight optimization.}
    \label{fig:codebleu_tcl}
    \vspace{-10pt}
    \Description{}
\end{figure}

As \Cref{tab:codebleu_comparison} presents, vanilla CodeBLEU (C1) incorrectly ranks \textit{Pythonic Tcl} higher than \textit{Standard Tcl}, because Pythonic constructs superficially align with the parser’s tokenization and syntax rules, even though they violate EDA semantics. To address this, we replace the Python keyword list with a curated Tcl/EDA command library covering 271 commands across four design servers. This module classifies commands by design stage and applies weighted penalties for invalid or misplaced options (e.g., misusing \texttt{compile\_ultra} in \Cref{fig:codebleu_tcl}, Weighted N-Gram Match). The resulting configuration (C2) raises the score of \textit{Standard Tcl} while lowering that of \textit{Pythonic Tcl}, demonstrating the benefit of domain-specific command awareness. However, improvements remain limited because the underlying parser still applies Pythonic structural rules.

To overcome this limitation, we develop a \textbf{Tcl-specific parser} that captures both the syntactic and semantic structures of EDA scripts. Unlike generic parsers, which fail to interpret Tcl’s command–argument patterns and variable substitution rules, our parser provides full grammar coverage and integrates two core modules.
The first, a \textbf{Data-Flow Graph Extractor}, builds command–variable dependency graphs to trace signal and configuration propagation across stages, enabling accurate modeling of EDA script logic and detection of semantic inconsistencies such as undefined or misused variables (\Cref{fig:codebleu_tcl}, \textit{Semantic Dataflow Match}).
The second, a \textbf{Syntax Matching Enhancement}, validates Tcl syntax at the line level in the absence of tree-sitter support. By excluding comments and blank lines, it focuses on meaningful command alignment while accommodating the irregular formatting and command chaining typical of industrial EDA scripts:

{\small
\[
\text{syntax\_match} = \frac{\sum_{i=1}^{n} \text{matching\_lines}_i}{\sum_{i=1}^{n} \text{total\_lines}_i}
\]}
Together, these modules enable the parser to recover both structural correctness and semantic intent, directly accounting for the significant improvement observed from C2 to C3 in \Cref{tab:codebleu_comparison}.

\begin{table}[t]
  \centering
  \footnotesize
  \caption{Evaluation scores across configurations. Keywords affect N-gram matching, and the parser enables AST and dataflow analysis. Pythonic Tcl mimics EDA commands using Python-like syntax.}
  \label{tab:codebleu_comparison}
  \vspace{-5pt}
  \begin{tabular}{lcc}
  \toprule \textbf{Config \{Keywords, Parser\}} 
      & \textbf{Pythonic Tcl} & \textbf{Standard Tcl} \\
  \midrule
  C1 \{Python,Python\} & 35.54 & 9.46 \\
  C2 \{TCL, Python\} & 33.24 & 12.24 \\
  C3 \{TCL, TCL\}    & 72.27 & \textbf{95.13} \\
  \bottomrule
  \end{tabular}
  \vspace{-10pt}
\end{table}

In summary, CodeBLEU-Tcl is the first evaluation framework that systematically extends CodeBLEU to the EDA domain, providing a reliable, domain-aware assessment of Tcl script quality in terms of both syntactic validity and tool-specific semantics.

\begin{table*}[ht]
    \centering
    \small
    \caption{Performance comparison across methods in terms of token usage, BLEU, and CodeBLEU scores. 
(S=Synthesis, P=Placement, C=Clock Tree Synthesis, R=Routing). Combinations such as S+P+C+R denote multi-stage workflows executed sequentially.}
    \vspace{-3pt}
    \setlength{\tabcolsep}{8pt}
    \renewcommand{\arraystretch}{1.1}
    \begin{tabular}{l ccc ccc ccc}
      \toprule
      \multirow{2}{*}{\textbf{Task}} 
        & \multicolumn{3}{c}{\textbf{Baseline 1}} 
        & \multicolumn{3}{c}{\textbf{Baseline 2}} 
        & \multicolumn{3}{c}{\textbf{\name}} \\
     \cmidrule(lr){2-4} \cmidrule(lr){5-7} \cmidrule(lr){8-10}
       & \textbf{Token} & \textbf{BLEU} & \textbf{CodeBLEU} 
       & \textbf{Token} & \textbf{BLEU} & \textbf{CodeBLEU} 
       & \textbf{Token} & \textbf{BLEU} & \textbf{CodeBLEU} \\
      \midrule
      Synthesis (S)     & 2532.64 & 0.450 & 10.720 
                  & 10376.39 & 98.382 & 98.836 
                  & \textbf{276.37}  & \textbf{98.413} & \textbf{98.852} \\
      Placement (P)     & 1781.46 & 0.030 & 5.775 
                  & 13340.68 & 98.800 & 99.004 
                  & \textbf{312.20}  & \textbf{98.899} & \textbf{99.035} \\
      CTS (C)     & 1478.48 & 1.782 & 17.685 
                  & 9116.78 & 96.902 & 97.878 
                  & \textbf{247.78}  & \textbf{97.284} & \textbf{98.084} \\
      Routing (R)     & 1384.49 & 0.388 & 7.942 
                  & 9464.82 & 100.000 & 100.000 
                  & \textbf{170.24}  & \textbf{100.000} & \textbf{100.000} \\
      \midrule
      S+P         & 2600.78 & 0.218 & 6.328 
                  & 15292.71 & 98.634 & 98.993 
                  & \textbf{460.66}  & \textbf{98.633} & \textbf{99.023} \\
      P+C         & 2220.78 & 0.991 & 10.093 
                  & 14035.13 & 97.703 & 98.370 
                  & \textbf{427.38}  & \textbf{97.703} & \textbf{98.474} \\
      C+R         & 1608.65 & 0.974 & 11.882 
                  & 10155.43 & 98.268 & 98.799 
                  & \textbf{280.74}  & \textbf{98.579} & \textbf{98.999} \\
      \midrule
      S+P+C       & 2478.90 & 0.467 & 6.929 
                  & 15985.71 & 97.952 & 98.544 
                  & \textbf{568.00}  & \textbf{97.952} & \textbf{98.650} \\
      P+C+R       & 2205.41 & 0.704 & 8.527 
                  & 15071.32 & 98.311 & 98.776 
                  & \textbf{452.04}  & \textbf{98.592} & \textbf{98.941} \\
      \midrule
      S+P+C+R     & 3059.42 & 0.408 & 5.743 
                  & 17017.06 & 98.485 & 98.902 
                  & \textbf{586.85}  & \textbf{98.585} & \textbf{98.950} \\
     \midrule
      Average     & 2075.06 & 0.641 & 9.961 
                  & 12869.52 & 98.324 & 98.810 
                  & \textbf{378.13}  & \textbf{98.564} & \textbf{98.901} \\
      \bottomrule
    \end{tabular}
    \vspace{-15pt}
    \label{tab:result}
\end{table*}

\section{Experiments} \label{sec:exp}

\subsection{Experimental Setup}
We evaluate the method on a diverse set of RTL designs in both Verilog and VHDL. The evaluation covers the four stages in \Cref{sec:mcp_server}, including logic synthesis using Synopsys Design Compiler \cite{synopsys_dc_user_guide}, and the physical design stages of placement, CTS, and routing using Cadence Innovus \cite{cadence_innovus_user_guide}. Both single-stage flows and multi-stage flows are considered. 
To support this evaluation, we generate 10k prompt–Tcl configuration pairs using our benchmark pipeline (\Cref{sec:benchmark}) and employ these prompts to produce Tcl scripts with three methods: two baselines and our proposed framework.

\noindent\textbf{Baselines.}  
We compare our method against two representative baselines. In both baselines, \textit{gpt-4o} is used to identify the target EDA stages and generate the corresponding Tcl scripts:

\begin{itemize}[leftmargin=*]
    \item \textit{\textbf{Baseline 1 (Direct Generation)}}: The LLM generates the Tcl script directly from the prompt with instructions that specify the intended tasks at each stage of the design flow.
    \item \textit{\textbf{Baseline 2 (In-Context Learning)}}: Prompts are augmented with template Tcl examples to guide LLM during generation.
    \item \name: Our proposed framework, consisting of the client model (Qwen3-0.6B finetuned with supervised training, see \Cref{sec:client}) and the server module described in \Cref{sec:mcp_server}.
\end{itemize}

\noindent\textbf{Evaluation Metrics.}  
The quality of the generated scripts is assessed using below complementary metrics:

\begin{itemize}[leftmargin=*]
    \item \textbf{\textit{CodeBLEU}} \cite{codebleu}: A syntactic and semantic-aware similarity metric for code generation, extended here to evaluate Tcl scripts on a 0–100 scale  (\Cref{sec:codebleu}).
    \item \textbf{\textit{BLEU}}\cite{bleu}: The n-gram–based text similarity metric on which CodeBLEU builds, serving as a supplementary metric to assess surface-level lexical similarity on a 0–100 scale.
    \item \textbf{\textit{Token Usage}}: The average number of tokens in both the input prompt and generated output is reported to evaluate generation efficiency and associated computational overhead.
    \item \textbf{\textit{Execution Success Rate}}: Generated Tcl scripts are executed with corresponding EDA tools to validate functional execution.
\end{itemize}
These metrics jointly capture fidelity (CodeBLEU/BLEU), efficiency (token usage), and practicality (success rate).

\subsection{Experiment Results}
Our evaluation across multiple EDA stages demonstrates significant improvements in Tcl script generation quality for our approach. Our approach consistently outperforms baseline methods across single and multi-stage, achieving superior CodeBLEU scores while maintaining efficient token usage.


\begin{table}[b]
    \vspace{-7pt}
    \centering
    \small
    \caption{CodeBLEU performance comparison for complete RTL-to-GDSII (S+P+C+R) workflow, SA for \name.}
    \vspace{-5pt}
    \label{tab:syn_pla_cts_rou}
    \begin{tabular}{cccccc}
    \toprule
    \textbf{Method} & \textbf{\scriptsize{CodeBLEU}} & \textbf{Syntax} & \textbf{Dataflow} & \textbf{W-Ngram} & \textbf{Ngram} \\
    \midrule
    Baseline1 & 5.743 & 18.844 & 0.283 & 2.156 & 0.408\\
    Baseline2 & 98.902 & 98.085 & 99.956 & 99.045 & 98.485\\
    \textbf{SA} & \textbf{98.950} & \textbf{98.085} & \textbf{99.980} & \textbf{99.142} & \textbf{98.585}\\
    \bottomrule
    \end{tabular}
    \label{tab:breakdown}
    \end{table} 

\subsubsection{CodeBLEU Score Analysis}
The CodeBLEU evaluation results are summarized in  \Cref{tab:result}. In single-stage, our approach achieves a CodeBLEU score of 98.99 on average, which is over 9$\times$ higher than Baseline1 (10.53) and 0.065\% higher than Baseline2 (98.93). 
Baseline1 performs poorly because it generates invalid commands without any pretrained knowledge or in-context learning. In contrast, our approach systematically instantiates parameters in pre-defined Tcl templates, ensuring validity and reliability in the generated scripts. 
For routing, both Baseline2 and our method achieve a perfect CodeBLEU score, as it does not contain parameters for extraction.

For multi-stage workflows, \name sustains robust performance, achieving CodeBLEU scores ranging from 98.474 to 99.023 for two-stage and from 98.650 to 98.941 for three-stage combinations. For the complete RTL-to-GDSII flow, detailed in \Cref{tab:breakdown}, ours achieves the highest overall CodeBLEU score of 98.950. 
Baseline methods degrade in quality when generating multiple scripts simultaneously, whereas \name benefits from the server's task decomposition and parameter validation mechanisms to guarantee correctness across stages. 
Compared to Baseline2 (98.902), the gain is modest. However, Baseline2 depends on a powerful closed-source model that is costly and raises privacy concerns. In contrast, \name merely relies on a locally fine-tuned 0.6B model, delivering comparable or better results with far greater efficiency and guaranteed data privacy. This highlights the practicality of our approach for real-world EDA automation.

\begin{figure}[t]
    \centering
    \includegraphics[width=1\linewidth]{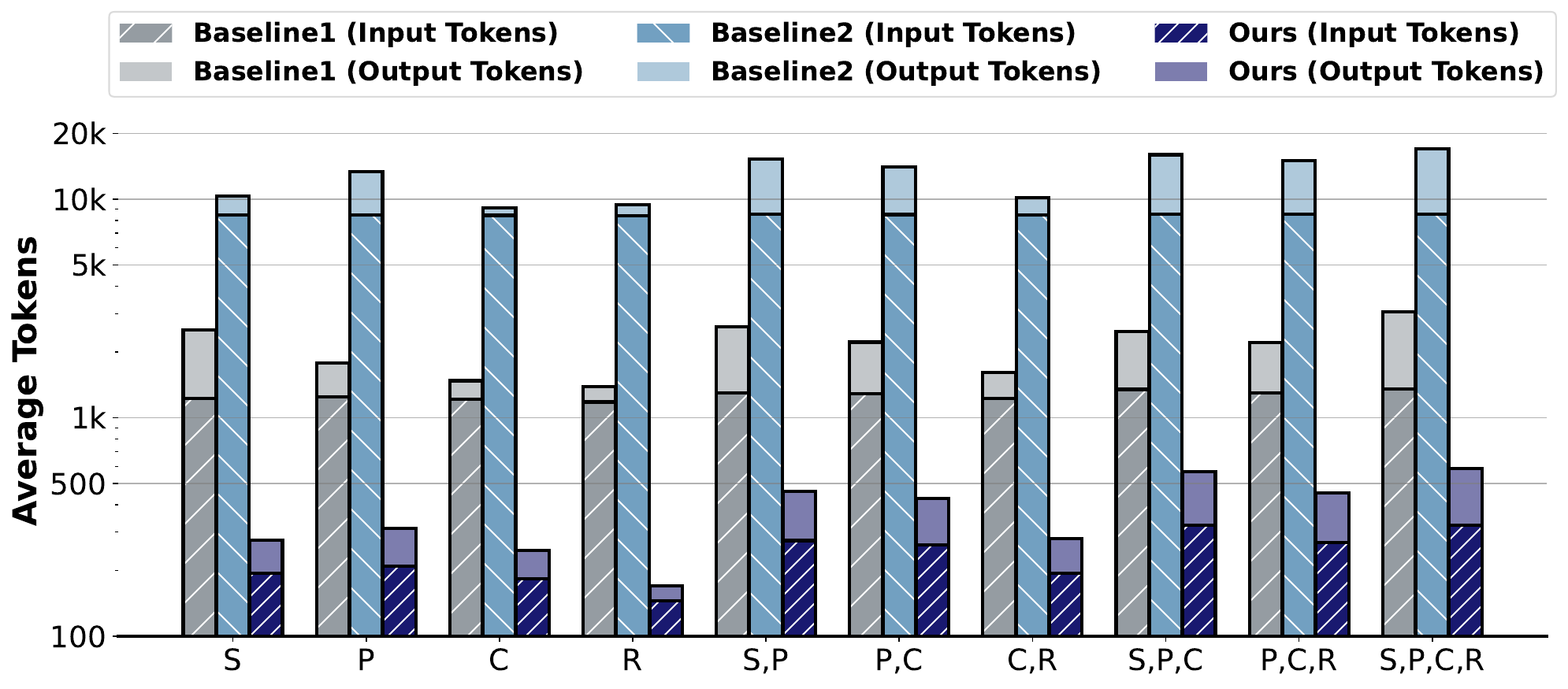}
    \caption{\textbf{Token usage across Benchmarks.} Our method achieves substantial token efficiency, using significantly fewer tokens than baseline2 while maintaining comparable or superior performance.}
    \label{fig:token_usage_comparison}
    \vspace{-10pt}
    \Description{}
\end{figure}

\subsubsection{Token Usage and Efficiency Analysis}
On average, as reported in \Cref{tab:result}, \name achieves the highest CodeBLEU score (98.901) with only 378 tokens, compared to 2075 tokens for Baseline1 (CodeBLEU 9.961) and 12,870 tokens for Baseline2 (CodeBLEU 98.810). 
This yields an 82\% and 97\% reduction over Baseline1 and Baseline2, respectively, while preserving or improving code quality.

\Cref{fig:token_usage_comparison} breaks down token usage across different flows. On the output side, Baseline1 and Baseline2 directly generate Tcl scripts, leading to verbose outputs. In contrast, \name only produces essential parameters, with the server handling script rendering. This design reduces output length by more than an order of magnitude. On the input side, Baseline2 requires in-context learning with example Tcl scripts, which substantially inflates the prompt size. By contrast, \name relies solely on user natural language requirements, avoiding redundant script templates. Consequently, it consistently reduces token usage across all benchmarks.

\begin{table}[t]
    \vspace{-7pt}
  \centering
  \footnotesize
  \caption{Comparison of token usage, success rate, and CodeBLEU score for RTL-to-GDSII (S+P+C+R) flow, SA for \name.}
  \vspace{-5pt}
  \begin{tabular}{l
      >{\centering\arraybackslash}p{0.9cm}
      >{\centering\arraybackslash}p{0.9cm}
      >{\centering\arraybackslash}p{0.9cm}
      >{\centering\arraybackslash}p{0.9cm}
      >{\centering\arraybackslash}p{0.9cm}
      >{\centering\arraybackslash}p{0.9cm}}
  \toprule
  \multirow{2}{*}{\textbf{Design}} 
      & \multicolumn{2}{c}{\textbf{Total Tokens}} 
      & \multicolumn{2}{c}{\textbf{CodeBLEU}} 
      & \multicolumn{2}{c}{\textbf{Success (\%)}} \\
  \cmidrule(lr){2-3} \cmidrule(lr){4-5} \cmidrule(lr){6-7}
      & \scriptsize{Baseline2} & \textbf{\scriptsize SA} 
      & \scriptsize{Baseline2} & \textbf{\scriptsize SA} 
      & \scriptsize{Baseline2} & \textbf{\scriptsize SA} \\
  \midrule
  b14      & 17011.55   & \textbf{578.25}   & 98.65 & \textbf{98.75} & 30.00  & \textbf{100.00} \\
  des      & 17023.85   & \textbf{588.65}   & 99.01 & \textbf{99.05} & 40.00 & \textbf{100.00} \\
  spi      & 17019.65   & \textbf{579.85}   & 99.01 & \textbf{99.01} & 35.00 & \textbf{100.00} \\
  s13207   & 17014.45   & \textbf{590.65}   & 98.82 & \textbf{98.93} & 45.00  & \textbf{100.00} \\
  s38584   & 17019.50   & \textbf{603.35}   & 98.89 & \textbf{98.95} & 35.00  & \textbf{100.00} \\
  \bottomrule
  \end{tabular}
  \label{tab:token_codebleu_success}
\end{table}

\subsubsection{Comprehensive Comparison with Baseline2}
\Cref{tab:token_codebleu_success} presents a detailed comparison with Baseline2 across five benchmark designs for the complete RTL-to-GDSII workflow (S+P+C+R). On average, \name requires only $588.15 \pm 8.60$ tokens per design, a 97\% reduction compared to Baseline2’s 17000+ tokens, while achieving comparable or slightly higher CodeBLEU scores (98.75–99.05 vs. 98.65–99.01). Although Baseline2 benefits from in-context Tcl templates and reaches near-perfect CodeBLEU, its generated scripts often contain subtle parameter errors. In the EDA domain, even minor inconsistencies can cause tool execution to fail, resulting in an average success rate of only 37\%. In contrast, \name enforces parameter validation and delegates script rendering to the MCP server, which ensures 100\% execution success across all benchmark designs. These findings underscore that in EDA automation, near-perfect code similarity does not guarantee executable results.

\begin{figure}[b]
    \centering
    \includegraphics[width=1\linewidth]{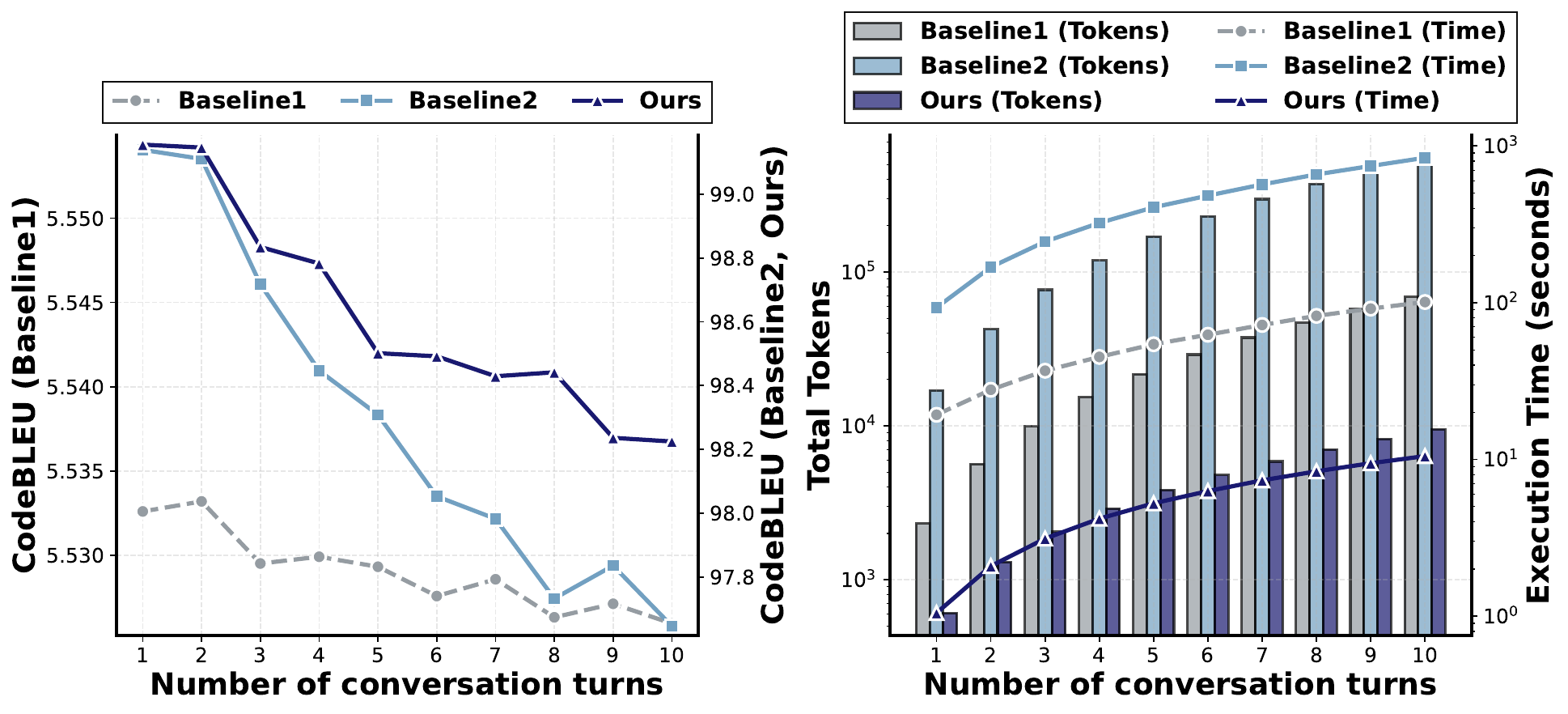}
    \caption{
    \textbf{CodeBLEU score, token usage, and execution time across number of conversation turns.} Our method exhibits substantial token and time efficiency while achieving better Tcl quality.}
    \label{fig:multi_turn}
    \vspace{-10pt}
    \Description{}
\end{figure}

\subsubsection{Performance Across Multi-Turn Conversations}
We further simulate a realistic multi-turn dialogue that reflects how users iteratively refine design parameters during RTL-to-GDSII automation. The initial query specifies the complete design flows, followed by successive prompts requesting parameter adjustments at each turn. Each subsequent LLM call received the full conversation history, including all prior user instructions and model responses, to maintain contextual continuity.
As shown in \Cref{fig:multi_turn}, \name, powered by a fine-tuned local LLM, achieves the lowest runtime and token usage while maintaining the highest CodeBLEU score. 
After ten turns, it requires merely $10.5s$ and $9.5k$ tokens, compared to Baseline 2’s $840s$ and $552k$ tokens. The widening CodeBLEU margin across turns highlights \name preserving Tcl-generation accuracy over long conversational sequences.

\subsection{AutoEDA}
This section illustrates how \name transforms a user’s high-level design intent into executable EDA commands.  
\begin{bluebox}{User Query}
\begin{lstlisting}[language=]
Synthesize design "b14" on FreePDK45 with fanout limit 4.74. Then run 
placement with a high level of effort for timing-driven global placer 
and medium wire length optimization effort level.
\end{lstlisting}
\end{bluebox}

\noindent Given the above description, the client transforms the request into a structured IR that specifies the design stages and parameters. IR is then wrapped in MCP format and passed to the server:

\begin{bluebox}{Intermediate Representation}
\begin{lstlisting}[language=Python]
"synth": {"design": "b14", "fanout_limit": 4.74},
"placement": {"design": "b14", "global_timing_effort": "high",
    "detail_wire_length_opt_effort": "medium"},
\end{lstlisting}
\end{bluebox}

\noindent The server then validates the IR, fills in defaults for unspecified inputs, and renders refined Tcl scripts for each stage. Finally, the executor runs these scripts on EDA tools, automatically generating the corresponding design artifacts and results.

\begin{bluebox}{Refined Tcl}
\begin{lstlisting}
####Tcl script for synthesis####
# as specified by the user
set MAX_FANOUT 4.74     
set TOP_NAME "b14"         
set_max_fanout $MAX_FANOUT $TOP_NAME
...
# not specified, use default values from server
set POWER_EFFORT "low" 
set AREA_EFFORT "medium" 
compile -power_effort $POWER_EFFORT -area_effort $AREA_EFFORT
...
####Tcl script for placement####
# as specified 
set PLACE_GLOBAL_TIMING_EFFORT "high" 
set PLACE_DETAIL_WIRE_LENGTH_OPT_EFFORT "medium" 
setPlaceMode -place_global_timing_effort $PLACE_GLOBAL_TIMING_EFFORT 
 -place_detail_wire_length_opt_effort $PLACE_DETAIL_WIRE_LENGTH_OPT_EFFORT 
placeDesign 
...
\end{lstlisting}
\end{bluebox}

% ...


\section{Conclusion}
We presented \name, a framework that leverages the Model Context Protocol to enable natural language control of RTL-to-GDSII flows. By combining MCP servers with locally fine-tuned models, \name addresses key challenges of reliability and privacy in LLM-driven EDA. The results demonstrate the potential of natural language–driven automation to make EDA flows more robust, efficient, and accessible. 

\newpage

\bibliographystyle{plain}
\bibliography{reference}

\end{document}